\documentclass[10pt,twocolumn,letterpaper]{article}

\usepackage{cvpr}
\usepackage{times}
\usepackage{epsfig}
\usepackage{graphicx}
\usepackage{amsmath}
\usepackage{amssymb}
\usepackage{url}
\usepackage{algorithm}
\usepackage{algorithmic}

\usepackage[british,UKenglish,USenglish,english,american]{babel}

\usepackage[bookmarks=false,colorlinks=true,linkcolor=black,citecolor=black,filecolor=black,urlcolor=black]{hyperref}

\cvprfinalcopy 

\def\cvprPaperID{****} 

\begin{document}

\title{Fast Guided Filter}

\author{Kaiming He \qquad Jian Sun\\
Microsoft\\
{\tt\small \{kahe,jiansun\}@microsoft.com}
}

\maketitle

\noindent\textbf{{Abstract}} \emph{The guided filter is a technique for edge-aware image filtering. Because of its nice visual quality, fast speed, and ease of implementation, the guided filter has witnessed various applications in real products, such as image editing apps in phones and stereo reconstruction, and has been included in official MATLAB and OpenCV. In this note, we remind that the guided filter can be simply sped up from $O(N)$ time to $O(N/s^2)$ time for a subsampling ratio $s$. In a variety of applications, this leads to a speedup of $>$$10\times$ with almost no visible degradation. We hope this acceleration will improve performance of current applications and further popularize this filter. Code is released.}

\section{Introduction}

The guided filter \cite{He2010,He2013} is one of several popular algorithms for edge-preserving smoothing\footnote{\url{en.wikipedia.org/wiki/Edge-preserving_smoothing}}. Its time complexity is $O(N)$ in the number of pixels $N$, independent of the filter size. The guided filter can effectively suppress gradient-reversal artifacts \cite{He2010} and produce visually pleasing edge profiles. Because of these and other properties, the guided filter has been included in official MATLAB 2014\footnote{\url{www.mathworks.com/help/images/ref/imguidedfilter.html}} and OpenCV 3.0\footnote{\url{docs.opencv.org/master/da/d17/group__ximgproc__filters.html}} and widely adopted in real products.

Despite its popularity and its various third-party implementations, we notice that a simple but significant speedup has not been exploited. This speedup strategy was briefly mentioned in \cite{He2013} for joint upsampling but not for other generic scenarios. This method subsamples the filtering input image and the guidance image, computes the local linear coefficients, and upsamples these coefficients. The upsampled coefficients are adopted on the \emph{original} guidance image to produce the output. This method reduces the time complexity from $O(N)$ to $O(N/s^2)$ for a subsampling ratio $s$. An actual speedup of $>$$10\times$ can be observed.

In this note, we revisit this speedup method by providing more technical details, visual examples, and publicly released code\footnote{\url{http://research.microsoft.com/en-us/um/people/kahe/eccv10}}.
This acceleration method is particularly favored for mega-pixel images, for which the filter size is usually set as proportional to the image size in practice. As such, a local window on the subsampled images can still provide enough pixels for computing local statistics. In our extensive real applications for image processing, we have found that this speedup method has almost no visible degradation.
Considering the growing usage of the guided filter in real products, we hope this simple speedup will improve the performance of these applications and further popularize this filtering technique.

\section{Method}

We denote the guidance image, filtering input image, and filtering output image as $I$, $p$ and $q$ respectively. The guided filter is driven by a local linear model:
\begin{equation}
q_{i}=a_{k}I_{i}+b_{k}, \forall i \in \omega_{k},\label{eq:linear}
\end{equation}
where $i$ is the index of a pixel, and $k$ is the index of a local square window $\omega$ with a radius $r$. Given the filtering input image $p$, minimizing the reconstruction error \cite{He2010} between $p$ and $q$ gives:
\begin{eqnarray}
a_{k}&=&\frac{\frac{1}{|\omega|}\sum_{i\in\omega_{k}}I_{i}p_{i}-\mu_{k}\bar{p}_{k}}{\sigma_{k}^{2}+\epsilon}\label{eq:a}\\
b_{k}&=&\bar{p}_{k}-a_{k}\mu_{k}\label{eq:b}.
\end{eqnarray}
where $\mu_{k}$ and $\sigma_{k}$ are the mean and variance of $I$ in the window $k$, and $\epsilon$ is a regularization parameter controlling the degree of smoothness.
The filtering output is computed by:
\begin{equation}
q_{i}=\bar{a}_{i}I_{i}+\bar{b}_{i}\label{eq:q},
\end{equation}
where $\bar{a}_{i}$ and $\bar{b}_{i}$ are the average of $a$ and $b$ respectively on the window $\omega_{i}$ centered at $i$.
The main computation is a series of box filters. Algorithm~\ref{alg:guidedfilter} shows the pseudo-code \cite{He2013} of the guided filter, where $f_\mathrm{mean}(\cdot, r)$ denotes a mean filter with a radius $r$.

In the above, $\bar{a}$ and $\bar{b}$ in Eqn.(\ref{eq:q}) are two smoothed maps, and the edges and structures in $q$ are mainly given by modulating the image $I$ (thus called \emph{guidance}). But the major computation of guided filter is for the smoothed maps of $\bar{a}$ and $\bar{b}$, which need not be performed in full-resolution.
Algorithm~\ref{alg:guidedfilter1} describes the subsampled version for \emph{Fast Guided Filter}. We subsample (nearest-neighbor or bilinear) the input $p$ and the guidance $I$ by a ratio $s$. All the box filters are performed on the low-resolution maps, which are the major computation of the guided filter. The two coefficient maps $\bar{a}$ and $\bar{b}$ are bilinearly upsampled to the original size. Finally, the output $q$ is still computed by $q=\bar{a}I+\bar{b}$. In this last step, the image $I$ is the full-resolution guidance that is \textbf{not} downsampled, and it will still faithfully guide the output.

\begin{figure}[t]
  \centering
  \includegraphics[width=1.0\linewidth]{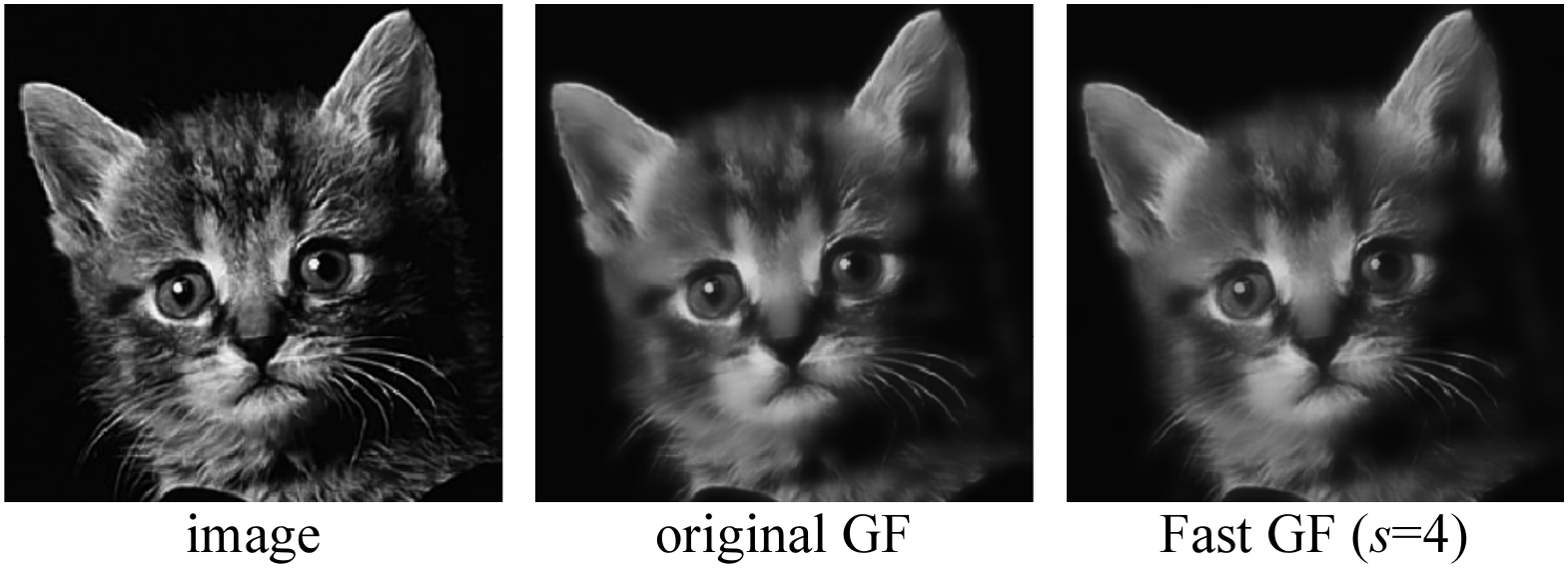}\\
  \caption{Edge-preserving smoothing. $r=4$, $\epsilon=0.2^2$. The subsampling ratio is $s=4$.}\label{fig:cat}
~\\
  \centering
  \includegraphics[width=1.0\linewidth]{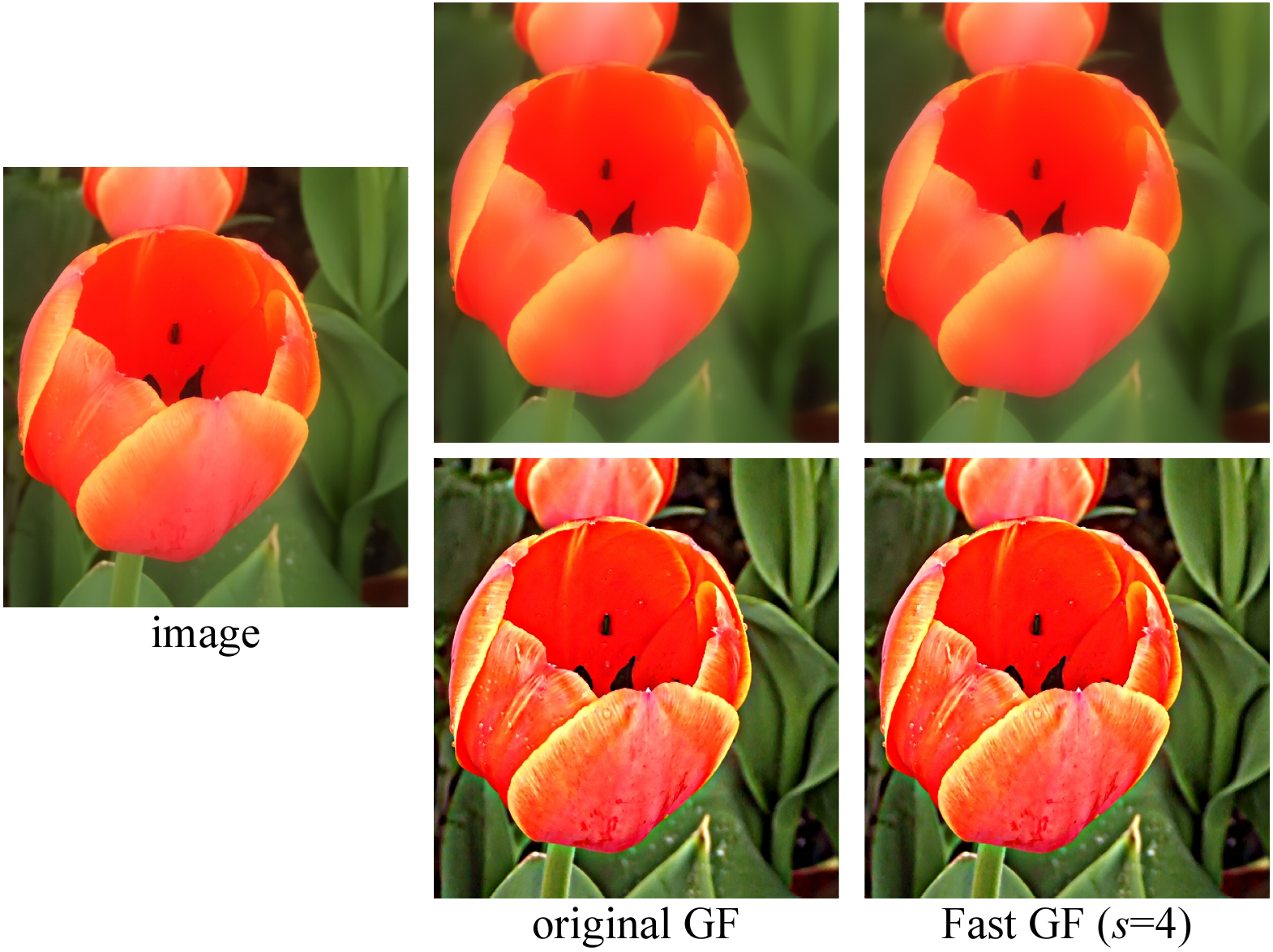}\\
  \caption{Detail enhancement. $r=16$, $\epsilon=0.1^2$. The subsampling ratio is $s=4$. Top: edge-preserving smoothed images. Bottom: enhanced images with $\times5$ detail.}\label{fig:tulips}
~\\
  \centering
  \includegraphics[width=1.0\linewidth]{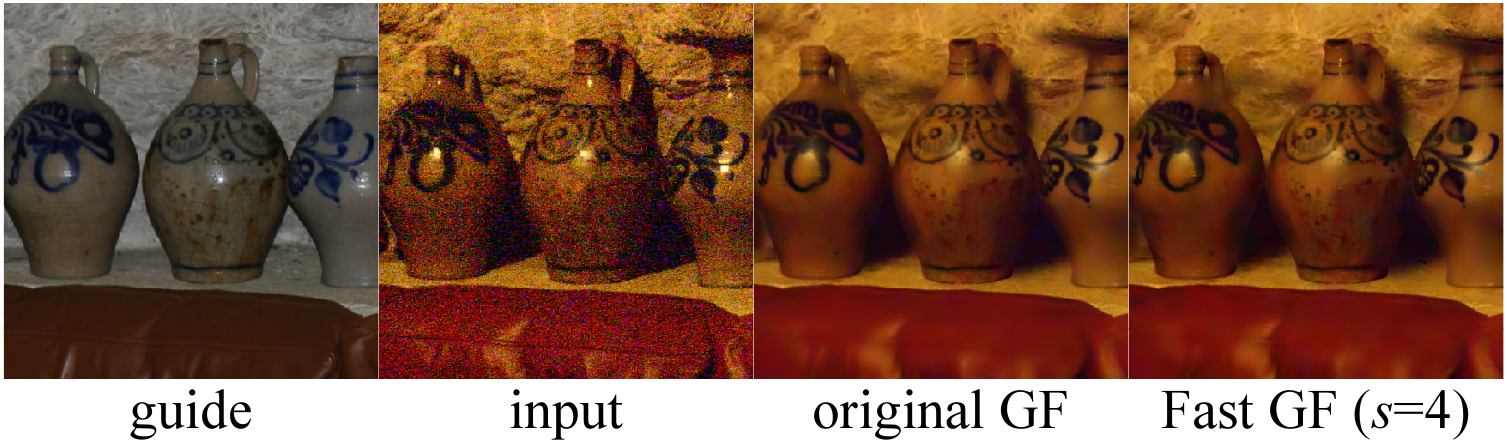}\\
  \caption{Flash/no-flash denoising. $r=8$, $\epsilon=0.02^2$. The subsampling ratio is $s=4$.}\label{fig:cave}
~\\
  \centering
  \includegraphics[width=1.0\linewidth]{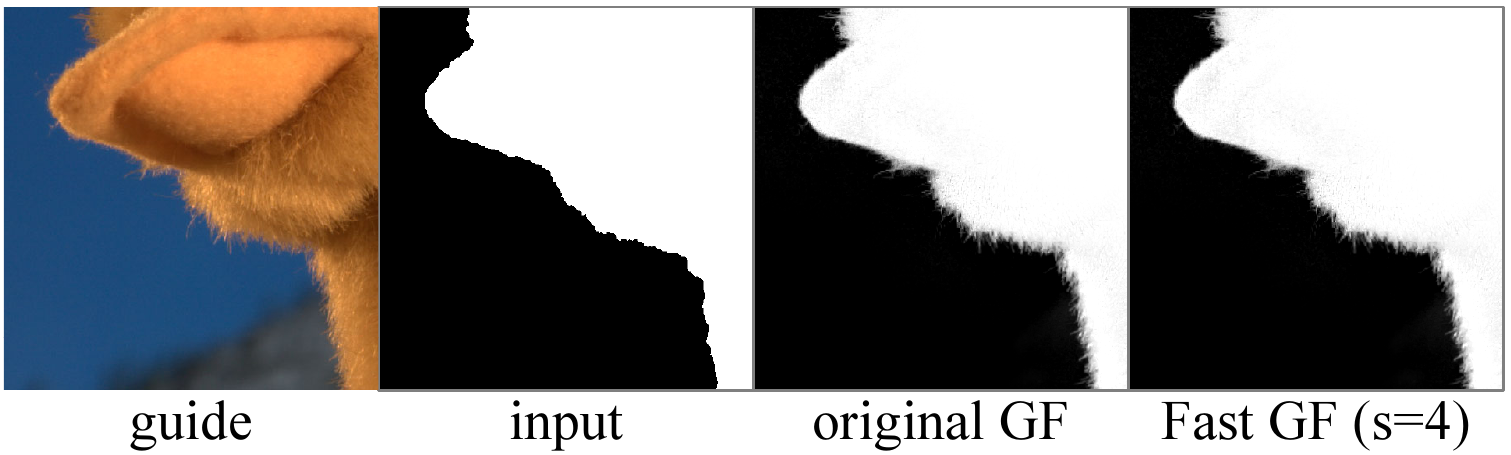}\\
  \caption{Guided feathering. $r=60$, $\epsilon=0.001^2$. The subsampling ratio is $s=4$.}\label{fig:feather}
\end{figure}

The computation of all box filters reduces from $O(N)$ complexity to $O(N/s^2)$. The last bilinear upsampling and output steps are $O(N)$ complex, but only take a small fraction of overall computation. In practice, we have observed a speedup of $>$$10\times$ when $s=4$ for both MATLAB and carefully optimized C++ implementation (depending on the number of channels).
Fig.~\ref{fig:cat}-\ref{fig:feather} show the visual results of using $s=4$ for various applications in \cite{He2010}.

\begin{algorithm}[t]
\small
\caption{\textbf{Guided Filter}. }
\begin{algorithmic}[1]\label{alg:guidedfilter}
\STATE $\mathrm{mean}_I=f_\mathrm{mean}(I, r)$\\
$\mathrm{mean}_p=f_\mathrm{mean}(p, r)$\\
$\mathrm{corr}_I=f_\mathrm{mean}(I.*I, r)$\\
$\mathrm{corr}_{Ip}=f_\mathrm{mean}(I.*p, r)$\\
\STATE $\mathrm{var}_I=\mathrm{corr}_I-\mathrm{mean}_I.*\mathrm{mean}_I$\\
$\mathrm{cov}_{Ip}=\mathrm{corr}_{Ip}-\mathrm{mean}_I.*\mathrm{mean}_p$\\
\STATE $a=\mathrm{cov}_{Ip}~./~(\mathrm{var}_I+\epsilon)$\\
$b=\mathrm{mean}_p-a.*\mathrm{mean}_I$\\
\STATE $\mathrm{mean}_a=f_\mathrm{mean}(a, r)$\\
$\mathrm{mean}_b=f_\mathrm{mean}(b, r)$\\
\STATE $q=\mathrm{mean}_a.*I+\mathrm{mean}_b$\\
\end{algorithmic}
\end{algorithm}

\begin{algorithm}[t]
\small
\caption{\textbf{Fast Guided Filter}. }
\begin{algorithmic}[1]\label{alg:guidedfilter1}
\STATE \underline{$I'=f_\mathrm{subsample}(I, s)$}\\
\underline{$p'=f_\mathrm{subsample}(p, s)$}\\
\underline{$r'=r/s$}\\
\STATE $\mathrm{mean}_I=f_\mathrm{mean}(I', r')$\\
$\mathrm{mean}_p=f_\mathrm{mean}(p', r')$\\
$\mathrm{corr}_I=f_\mathrm{mean}(I'.*I', r')$\\
$\mathrm{corr}_{Ip}=f_\mathrm{mean}(I'.*p', r')$\\
\STATE $\mathrm{var}_I=\mathrm{corr}_I-\mathrm{mean}_I.*\mathrm{mean}_I$\\
$\mathrm{cov}_{Ip}=\mathrm{corr}_{Ip}-\mathrm{mean}_I.*\mathrm{mean}_p$\\
\STATE $a=\mathrm{cov}_{Ip}~./~(\mathrm{var}_I+\epsilon)$\\
$b=\mathrm{mean}_p-a.*\mathrm{mean}_I$\\
\STATE $\mathrm{mean}_a=f_\mathrm{mean}(a, r')$\\
$\mathrm{mean}_b=f_\mathrm{mean}(b, r')$\\
\STATE \underline{$\mathrm{mean}_a=f_\mathrm{upsample}(\mathrm{mean}_a, s)$}\\
\underline{$\mathrm{mean}_b=f_\mathrm{upsample}(\mathrm{mean}_b, s)$}\\
\STATE $q=\mathrm{mean}_a.*I+\mathrm{mean}_b$\\
\end{algorithmic}
\end{algorithm}

{\small
\bibliographystyle{ieee}
\bibliography{fgf}
}

\end{document}